\documentclass[runningheads]{llncs}
\usepackage{draftwatermark}
\SetWatermarkText{Preprint}
\SetWatermarkScale{4}
\SetWatermarkColor[gray]{0.90}
\usepackage{graphicx}
\usepackage{caption}
\usepackage{subcaption}

\usepackage{algorithm}
\usepackage{algorithmicx}
\usepackage[noend]{algpseudocode}

\usepackage{booktabs}
\usepackage{multirow}

\makeatletter
\newcommand{\HEADER}[1]{\ALC@it\underline{\textsc{#1}}\begin{ALC@g}}
\newcommand{\ENDHEADER}{\end{ALC@g}}
\makeatother

\usepackage{amsmath,amsfonts}

\usepackage{nicefrac}

\usepackage[normalem]{ulem}
\usepackage{xcolor}
\usepackage{soul}

\usepackage{xspace}
\usepackage{hyperref}

\makeatletter
\DeclareRobustCommand\onedot{\futurelet\@let@token\@onedot}
\def\@onedot{\ifx\@let@token.\else.\null\fi\xspace}

\def\eg{\emph{e.g}\onedot} 
\def\ie{\emph{i.e}\onedot}

\makeatother

\begin{document}
\title{Integrating Hyperparameter Search into Model-Free AutoML with Context-Free Grammars}

\titlerunning{Integrating Hyperparameter Search into GramML}
\author{Hernán Ceferino Vázquez \and
Jorge Sanchez \and
Rafael Carrascosa}
\authorrunning{H. C. Vazquez et al.}
\institute{MercadoLibre Inc.\\
\email{\{hernan.vazquez,jorge.sanchez,\\rafael.carrascosa\}@mercadolibre.com}
}

\maketitle              % typeset the header of the contribution
\begin{abstract}

Automated Machine Learning (AutoML) has become increasingly popular in recent years due to its ability to reduce the amount of time and expertise required to design and develop machine learning systems. This is very important for the practice of machine learning, as it allows building strong baselines quickly, improving the efficiency of the data scientists, and reducing the time to production. However, despite the advantages of AutoML, it faces several challenges, such as defining the solutions space and exploring it efficiently. Recently, some approaches have been shown to be able to do it using tree-based search algorithms and context-free grammars. In particular, GramML presents a model-free reinforcement learning approach that leverages pipeline configuration grammars and operates using Monte Carlo tree search. However, one of the limitations of GramML is that it uses default hyperparameters,  limiting the search problem to finding optimal pipeline structures for the available data preprocessors and models. In this work, we propose an extension to GramML that supports larger search spaces including hyperparameter search. We evaluated the approach using an OpenML benchmark and found significant improvements compared to other state-of-the-art techniques.

\end{abstract}

\section{Introduction}

The practice of Machine Learning (ML) involves making decisions to find the best solution for a problem among a myriad of candidates. Machine learning practitioners often find themselves in a process of trial and error in order to determine optimal processing pipelines (data preprocessing, model architectures, etc.) and their configurations (hyperparameters). These decisions go hand in hand with the many constraints imposed by the actual application domain and have a great impact on the final system performance. These constraints can be either intrinsic, such as working with heterogeneous data, extrinsic, such as requiring using limited computational resources and/or short development cycles, or both. Solving real-world problems by means of ML is a time-consuming and difficult process.

%In this scenario, Automated Machine Learning (AutoML) emerges as an area of main interest since it focuses on developing methods to automate the generation of machine learning models \cite{vazquez2023general}. Due to its ability to reduce the amount of time and expertise required to design and develop machine learning models AutoML has become increasingly popular in recent years.

In this context, the development of methods and techniques that allow for the automated generation of machine learning solutions \cite{vazquez2023general} is a topic of great practical importance, as they help reduce the time and expertise required for the design and evaluation of a broad range of ML-based products and applications.

%Despite the benefits that this brings to the practice of machine learning, AutoML has several challenges. Two of the main ones are: how to define the solution search space in a way that is both comprehensive and computationally feasible; and how to explore the search space efficiently in terms of time and computational resources.

Besides the benefits it brings to the practice of machine learning, AutoML poses several challenges. In our case, these challenges are how to define the solution space (the set of possible pipeline configurations, i.e. pipeline structure and its hyperparameters) in a way that is both comprehensive and computationally feasible, and how to explore it efficiently both in terms of time and resources.

%\comment{a lo que resta de la sección lo organizaría en los siguientes párrafos:
%\begin{itemize}
%    \item model-based vs model-free: contar que la tendencia es usar modelos model-based pero que tienen el problema de uqe requieren metafeatures, etc. Contar que una alternativa son los model-free y dar referencias (solo menciones, no discutirlos) más alla de gramml. 
%    \item entre los model-free una estrategia interesante es el uso de gramáticas y justificarlo en la flexibilidad de modelado (mantenible, adaptable, escalable) y lo amigables que son para búsqueda (tree). Decir acá que tomamos como punto de partida el modelo GramML propuesto por nos. Decir que una limitación de GramML es que no se expanden los hparam y que el algo de búsqueda no es escalable
%   \item presentar la propuesta del paper 
%\end{itemize}
%}

%\cefe{Many techniques have been developed over the years to address this problem. Most of them relies on models to guide the search space exploration (model-based). However, these approaches highly depend on how the space and solutions are represented as metafeatures and of the meta-model as well as its hyperparameters. An alternative to model-based approaches are model-free approaches. Model-free approaches uses another kind of strategies to guide the search process in which evolutionary approaches and tree-based techniques are the main ones.} 

Many techniques have been developed over the years to address these problems, most of which rely on parametric models to guide the exploration of the search space. However, such model-based solutions require a characterization of configuration space and its possible solutions via meta-features, which are not always easy to define and evolve over time, adding an extra (meta) learning step as well as an additional layer of hyperparameters (those of the meta-model) \cite{vanschoren2019meta}. An alternative to this family of models corresponds to the model-free counterpart. These approaches rely on different strategies to guide the search stage, in which evolutionary \cite{olson2016tpot,evans2020adaptive}and tree-based techniques \cite{rakotoarison2018automl,marinescu2021searching} appear as the most prominent ones.

%\cefe{In recent years there has been a growing interest in tree-based strategies due to the synergy they offer with context-free grammars \cite{marinescu2021searching,drori2019automatic,vazquez2022gramml}. Grammars provide a way to specify a set of valid models, and can be used to control the complexity of the search space. In addition, grammars provide a flexible and extensible description of the configuration space and allow for tree-like navigation because the parse-tree.}

A particularly interesting approach consists of using context-free grammars for the search space specification and an efficient search strategy over the production tree \cite{waddle1990production} derived from it \cite{marinescu2021searching,drori2019automatic,vazquez2022gramml}. 
%
%\cefe{A specific algorithm that has been used in this context is Monte Carlo Tree Search (MCTS), which allows for a more efficient exploration of the space \cite{rakotoarison2018automl}. In particular GramML \cite{vazquez2022gramml}, present a simple yet effective model-free reinforcement learning approach based on an adaptation of the MCTS for trees and context-free grammars that show superior performance compared to the state-of-the-art. However, one of the limitations of GramML is that it uses default hyperparameters, thus reducing the search problem to finding good pipeline architectures for the different data preprocessors and models available.}
%
A specific algorithm that has been used in this case is Monte Carlo Tree Search (MCTS), which allows for a more efficient exploration of the search space \cite{rakotoarison2018automl}. In particular, \cite{vazquez2022gramml} proposes GramML, a simple yet effective model-free reinforcement learning approach based on an adaptation of the MCTS for trees and context-free grammars that show superior performance compared to the state-of-the-art. However, one of the limitations of their model is that it disregards hyperparameters, leaving them to their default values. This effectively reduces the search space pipeline architectures consisting of data preprocessors and generic model configurations.

%\cefe{In this work, we propose an extension to GramML that supports larger search spaces and in particular hyperparameter search. This extension is mainly based on a modification in the MCTS algorithm which incorporates pruning and the use of non-parametric selection policies.}

In this work, we propose an extension to GramML that supports larger search spaces by incorporating hyperparameters into the model. We extend the basic MCTS algorithm to account for this increase in complexity, by incorporating a pruning strategy and the use of non-parametric selection policies. In addition, we run experiments to evaluate the performance of our approach and compare it with state-of-the-art techniques. The results show that our method significantly improves performance and demonstrates the effectiveness of incorporating hyperparameter search into grammar-based AutoML.

\section{Related Work}

This section provides a succinct overview of prior studies related to AutoML. The main representative for solving this class of problems is AutoSklearn \cite{feurer2015efficient}, which is based on the Sequential Model-Based Algorithm Configuration (SMAC) algorithm  \cite{hutter2011sequential} and incorporates Bayesian optimization (BO) and surrogate models to determine optimal machine learning pipelines. In AutoSklearn hyperparameters and pipeline selection problem are tackled as a single, structured, joint optimization problem named CASH (Combined Algorithm Selection and Hyperparameter optimization \cite{thornton2013auto}). Recent research has produced model-based approaches that exhibit comparable performance to AutoSklearn. For instance, MOSAIC \cite{rakotoarison2018automl} leverages a surrogate model formulation in conjunction with a Monte Carlo Tree Search (MCTS) to search the hypothesis space. MOSAIC tackle both pipeline selection and hyperparameter optimization problems using a coupled hybrid strategy: MCTS is used for the first and Bayesian optimization for the latter, where the coupling is enforced by using a surrogate model. AlphaD3M \cite{drori2021alphad3m} also builds on MCTS and uses a neural network as surrogate model, considering only partial pipelines rather than whole configurations. The difference with MOSAIC is that the pipeline selection is decoupled from the hyperparameter search, it first looks for the best pipelines and then performs hyperparameter optimization of the best pipelines using SMAC. \cite{drori2019automatic} studied the use of AlphaD3M in combination with a grammar, relying on pre-trained models for guidance during the search. Another approach, PIPER \cite{marinescu2021searching}, utilizes a context-free grammar and a greedy heuristic, demonstrating competitive results relative to other techniques. In PIPER, the best pipeline structure is found and then its hyperparameters are searched with a standard CASH optimizer. Other approaches include the use of genetic programming \cite{olson2016tpot,de2017recipe} and hierarchical planning \cite{mohr2018ml,katz2020exploring} techniques.

\section{Background}

%\comment{hay que explicar conceptualmente la gramática y cómo el problema de automl se transforma en uno de búsqueda en un árbol.} \cefe{Creo que algo asi queda bien. No se si es necesario un diagrama}
A formal grammar is a collection of symbols and grammar rules that describe how strings are formed in a specific formal language \cite{segovia2017generating}. There are various types of grammar, including context-free grammars \cite{chomsky2009syntactic}. In context-free grammars, the grammar rules consist of two components: the left-hand side of a grammar rule is always a single non-terminal symbol. These symbols never appear in the strings belonging to the language. The right-hand side contains a combination of terminal and non-terminal symbols that are equivalent to the left-hand side. Therefore, a non-terminal symbol can be substituted with any of the options specified on the right-hand side.

Once specified, the grammar provide a set of rules by which non-terminal symbols can be substituted to form \emph{strings}. These strings belong to the language represented by the grammar. This generation process (production task) is performed by starting from an initial non-terminal symbol and replacing it recursively using the grammar rules available. An important property, in this case, is that the set of rules can be expressed in the form of a tree and the generation of any given production can be seen as a path from the root to a particular leaf of this tree. The generation process can be thus seen as a search problem over the tree induced by the grammar.

In the context of the AutoML problem, the grammar is used to encode the different components and their combination within a desired pipeline. For instance, we can set a grammar that accounts for different data preprocessing strategies, like \emph{PREPROCESSING := "NumericalImputation" or SCALING := "MinMaxScaler" or "StandardScaler"} or the type of weight regularization used during learning, as \emph{PENALTY := "l1" or "l2"}. Finding optimal ML pipelines corresponds to searching for a path from the root (the initial symbol of the grammar) to a leaf of the tree (the last rule evaluated) that performs best in the context of a given ML problem. Interestingly, we could also encode hyperparameter decisions within the same grammar, \eg by listing the possible values (or ranges) that a particular parameter can take. By doing so, we grow the tree size (configuration space of all possible ML pipelines and their configurations) exponentially.

%\comment{[CHECK]} \cefe{as the number of decisions to take increases}. \cefe{This is because each decision adds another dimension to the space. For example, consider a model with two hyperparameters, each with a range of values from 1 to 10. The total number of possible combinations (represented as leaves on the tree) of hyperparameter values is 100 (10 x 10, or $10^2$). However, if a third hyperparameter is added with a range of values from 1 to 10, the total number of possible combinations increases to 1000 (10 x 10 x 10, or $10^3$). Roughly, running a brute force algorithm over the entire space has a time complexity of $O(v^d)$. Where, $d$ is the number of decisions ($=3$ hyperparameters in the example) and $v$ is the number of values of each one ($=10$ possibilities in example).}

From the above, a key component of grammar-based approaches to AutoML is the search algorithm, as it has to be efficient and scale gracefully with the size of the tree. In the literature, one of the most reliable alternatives is the Monte Carlo Tree Search (MCTS) algorithm \cite{kocsis2006bandit}. MCTS is a search algorithm that seeks to expand the search tree by using random sampling heuristics. It consists of four steps: \emph{selection}, \emph{expansion}, \emph{simulation} and \emph{backpropagation}. During the selection step, the algorithm navigates the search tree to reach a node that has not yet been explored. Selection always starts from the root node and, at each level, selects the next node to visit according to a given \emph{selection policy}. The expansion, simulation, and backpropagation steps ensure that the tree is expanded according to the rules available at each node and, once a terminal node is reached, that the information about the reliability of the results obtained at this terminal state is propagated back to all the nodes along the path connecting this node to the root. As a reference, Algorithm \ref{algo:basic_mcts} illustrates the general form of the MCTS algorithm.

%\subsection{Monte Carlo Tree Search}
%
%Monte Carlo Tree Search (MCTS) is a reinforcement learning approach that uses heuristic exploration to construct its search tree, which is composed of nodes, each node having multiple child nodes that correspond to the available actions of the agent. MCTS algorithm involves the following steps: 
%
%\begin{enumerate}
%    \item \emph{Selection}. The agent navigates the search tree to reach a node that has not yet explored all \suggest{of }{}its child nodes. Selection always starts from the root node and, at each level, selects the next node according to the selection policy, also referred to as tree policy. 
%    \item \emph{Expansion}. Once a node with unexplored children is reached, the agent creates a new random child node and simulates the environment for that node. In some problems it is extremely rare to reach terminal states, but if it does, the current iteration skips directly to backpropagation.
%    \item \emph{Simulation}. The agent uses Monte Carlo simulations to follow a random policy until the environment reaches a terminal state. Once the final state is reached, results are computed.
%    \item \emph{Backpropagation}. The agent uses the scores received in the terminal state to propagate them to all nodes along the path from leaf to root. 
%\end{enumerate}
%
%These steps are repeated a set number of times to incrementally build the search tree.
%

\begin{algorithm}
\caption{General MCTS approach}
\begin{algorithmic}[1]
\Function{Search}{$s_0$}
\State create root node $a_0$ with state $s_0$
\While{within budget}
\State $a_i \gets \text{Selection}(a_0, SelectionPolicy)$
\State $a_i \gets \text{Expansion}(a_i)$
\State $\Delta \gets \text{Simulation}(s(a_i))$
\State $\text{Backpropagation}(a_i, \Delta)$
\EndWhile
\Statex \Return{$\text{BestChild}(a_0, SelectionPolicy)$}
\EndFunction
\end{algorithmic}
\label{algo:basic_mcts}
\end{algorithm}

In general, MCTS is a powerful method that allows to explore the search tree in a systematic and efficient way and can be used for AutoML. However, balancing the exploration and exploitation trade-off depends on find best selection policy to use. Furthermore, when the tree grows up exponentially as in the case of hyper-parameter search problem, finding good solutions in an acceptable time can be challenging.

\section{Hyperparameter Search in Grammar-Based AutoML}

%Efficiently exploring the space defined by AutoML grammars can be accomplished using a model-free MCTS. For this we extended the GraMML approach by adding hyperparameters to the grammar and implementing a pruning mechanism to reduce space after each iteration. In addition, we implement different selection policies. A detailed description of GramML's extended approach to support hyperparameter search is described in the following sections.

%\subsection{Model-Free MCTS for AutoML with Hyperparams}

%In GramML, the objective is to generate productions of the grammar (i.e. pipeline configurations) using the available production rules that encode the choices for a particular non-terminal symbol. In the case of hyperparameters, the options are included as part of the grammar. For continuous numerical hyperparameters, we sample within defined ranges representative values of each one. In the case of categorical or discrete hyperparameters we incorporate the options directly into the grammar.

In grammar-based AutoML, the objective is to generate pipeline configurations using the available production rules that encode the choices for a particular non-terminal symbol. In our case, we extend the GramML\footnote{The Extended GramML source code is available at: \href{https://github.com/mercadolibre/fury_gramml-with-hyperparams-search}{github.com/mercadolibre/fury\_gramml-with-hyperparams-search}.} formulation of \cite{vazquez2022gramml} to account for the hyperparameters associated with the different ML components that compose an ML pipeline. For continuous numerical hyperparameters, we use the same sampling intervals as AutoSklearn for a fairer comparison. In the case of categorical or discrete parameters, we incorporate their values directly into the grammar.

Using MCTS the grammar is traversed generating productions and looking for the best pipeline configurations. The algorithm can be stopped at any point, and the current best pipeline configuration is returned. The longer the algorithm runs, the more productions will be generated and evaluated, and the more likely that the recommended pipeline is indeed the optimal one. The process of generating grammar productions and evaluating pipeline configurations consists of the traditional four steps of MCTS; however, it has some considerations given the nature of the problem. In our setting, the only valid configurations are leaf nodes, as they contain the complete configuration object. Any attempt to execute a partial configuration would result in an error. Additionally, each iteration of the algorithm (commonly referred to as an episode in RL) ends in a leaf node or terminal state. These differences lead to some changes in the original algorithm.

\begin{algorithm}
\caption{MCTS for AutoML}
\begin{algorithmic}[1]
\Function{Search}{$s_0$}
\State $a_i \leftarrow a_0$
\While{used budget $<$ budget $\And$ $a_i$ is not Terminal}
\State $a_i \gets \text{Selection}(a_0, SelectionPolicy)$
\If{$a_i$ is not Terminal}\State $a_i \gets \text{Expansion}(a_i)$\State $\Delta \gets \text{Simulation}(s(a_i))$
\Else
\State $\Delta \gets \text{GetReward}(s(a_i))$
\EndIf
\State $\text{Backpropagation}(a_i, \Delta)$
\EndWhile
\State {$p_i^* \leftarrow \text{BestLeaf}(a_0, SelectionPolicy)$}
\State {$\text{PruneTree}(p_i^*)$}
\State \Return{$p_i^*$}
\EndFunction
\end{algorithmic}\label{alg:search}
\end{algorithm}

%Algorithm~\ref{alg:search} shows the search function of MCTS for AutoML. The search function is called by the main program until the budget is exhausted. The budget can be a specified number of iterations or a time duration. Each call to the search function returns a pipeline configuration. The function is executed until either a pipeline configuration is found or the budget is exhausted.
Algorithm~\ref{alg:search} illustrates an adaptation of the MCTS algorithm for AutoML. The \textsc{Search} function is called by the main program until a budget is exhausted. The budget can be specified in terms of the number of iterations or running time. Each call to the \textsc{Search} function returns a pipeline configuration. This function is executed until either a pipeline configuration is found or the budget is exhausted.

In each step of the algorithm, if the selected node is not terminal, the search function performs the expansion and simulation steps. In the expansion step, the tree is navigated from the selected node to find a node that has not been expanded. Once found, it is expanded, and the algorithm continues. In the simulation step, the algorithm randomly selects a path from the selected node to a leaf. Upon reaching a leaf, a reward function is computed and returned. In the case of the selected node being terminal, this reward is used for backpropagation. Finally, the selected leaf is pruned from the tree. If all leaves of a branch are pruned, then the branch is also pruned.

%\subsection{Selection and Backpropagation}

%The selection and backpropagation steps depend on the selection policy used. Selection policies are important as they guide the search process of the algorithm in the tree, regulating the trade-off between exploration and exploitation. The selection function for each policy used, along with the backpropagation algorithm, is described below.
Selection and backpropagation steps depend on the actual selection policy. Selection policies are important as they guide the search, regulating the trade-off between exploration and exploitation. In what follows, we present three different alternatives to this combination that appear as particularly suited to our problem.

\paragraph{Upper Confidence Bound applied to Trees (UTC). } UTC \cite{kocsis2006bandit} is a standard algorithm that seeks to establish a balance between exploration and exploitation to find the best option among a set of candidates. It uses a constant C as the only parameter that regulates this trade-off. UCT is based on the computation of two statistics, a cumulative sum of the rewards obtained and the number of visits for each node. base on these values, we compute a score for each node as:
 \begin{equation}
 v(a_i) = \frac{reward(a_i)}{visits(a_i)} + C \sqrt \frac{\log{\left(visits(a_i)\right)}}{visits(parent(a_i)))},
 \label{equ:valueUCT}
\end{equation}

The selection policy corresponds to choosing the node with the highest score. Algorithm~\ref{alg:backprop_UCT} shows the process by which rewards and visits are backpropagated from a leaf up to the root.
\begin{algorithm}
\caption{Backpropagation for UCT}
\label{alg:backprop_UCT}
\begin{algorithmic}[1]
\Function{Backpropagation}{$a$, $\Delta$}
\State $a_i \leftarrow a$ \algorithmiccomment{a is a leaf node}
\While{$a_i$ is not Null}
    \State $reward_i \leftarrow reward_i + \Delta$
    \State $visits_i \leftarrow visits_i + 1$
%\State $a_i \leftarrow a_i$ parent  //  \textit{if $a_i$ is the root node parent is Null}
\State $a_i \leftarrow parent(a_i)$ \algorithmiccomment{if $a_i$ is the root node parent is Null}
\EndWhile
\EndFunction
\end{algorithmic}
\end{algorithm}

%Using rewards and visits, we calculate the values of the node $v(a_i)$ as shown in equation \ref{equ:valueUCT}.
%
%\begin{equation}
%a^*_i = \underset{i\in I}{\mathrm{arg\,max}}(v(a_i)),\label{equ:armaxUCT}
%\end{equation}
% where $I$ is the set of all child nodes of the parent of $a_i$ and $a^*_i$ is its best child.

\paragraph{Bootstrap Thompson Sampling (BTS).} In Thompson Sampling (TS) the basic idea is to select actions stochastically \cite{bai2013bayesian}. TS requires being able to compute the exact posterior \cite{chapelle2011empirical}. In cases where this is not possible, the posterior distribution can be approximated using bootstrapping \cite{efron2012bayesian}. This method is known as Bootstrap Thompson Sampling (BTS) \cite{eckles2014thompson}. In BTS the bootstrap distribution is a collection of bootstrap replicates $j \in \{1, ..., J\}$, where $J$ is a tunable hyperparameter that balances the trade-off between exploration and exploitation. A smaller $J$ value results in a more greedy approach, while a larger $J$ increases exploration, but has a higher computational cost \cite{eckles2014thompson}. Each replicate $j$ has some parameters $\theta$, which are used to estimate $j$'s expected utility given some prior distribution $P(\theta)$. At decision time, the bootstrap distribution for each node is sampled, and the child with the highest expected utility is selected. During backpropagation, the distribution parameters are updated by simulating a coin flip for each replicate $j$. In our scenario, we use a Normal distribution as prior and two parameters $\alpha$ y $\beta$ (by consistency to other works \cite{hayes2021distributional}.) If the coin flip comes up heads, the $\alpha$ and $\beta$ parameters for $j$ are re-weighted by adding the observed reward to $\alpha_j$ and the value $1$ to $\beta_j$.
Algorithm~\ref{alg:backprop_BTS} illustrates the backpropagation step for the BTS strategy.
\begin{algorithm}
\caption{Backpropagation for BTS}
\label{alg:backprop_BTS}
\begin{algorithmic}[1]
\Function{Backpropagation}{$a$, $\Delta$}
\State $a_i \leftarrow a$ \algorithmiccomment{a is a leaf node}
\While{$a_i$ is not Null}
\For{$j \in J$}
    \State sample $d_{j}$ from Bernoulli(\nicefrac{1}{2})
    \If{$d_{j} = 1$}
        \State $\alpha_{a_i} \leftarrow \alpha_{a_i} + \Delta$
        \State $\beta_{a_i} \leftarrow \beta_{a_i} + 1$
    \EndIf
\EndFor
%\State $a_i \leftarrow a_i$ parent \algorithmiccomment{if $a_i$ is the root node parent is Null}
\State $a_i \leftarrow parent(a_i)$ \algorithmiccomment{if $a_i$ is the root node parent is Null}
\EndWhile
\EndFunction
\end{algorithmic}
\end{algorithm}

%For BTS we decide for and stochastic selection policy. After computing the value of each node, it will have a probability of being selected $p(a_i)$ proportional to its value $v(a_i)$.

%\begin{equation}
%v(a_i)=E(\alpha_i,\beta_i)= \frac{\alpha_i}{\alpha_i+\beta_i}\label{equ:valueBTS}
%\end{equation}

%\begin{equation}
%p(a_i, v(a_i)) = \frac{v(a_i)}{\sum v(a_c)}\label{equ:armaxBTS}
%\end{equation}

%where $\sum v(a c)$ is the sum of the values of all nodes of the %same parent as $a_i$.

Regarding the selection policy, we estimate the value of the nodes as $\alpha_{a_i}/\beta_{a_i}$, which represents the largest point estimate of our prior (the mean). Afterwards, we select a child with a probability proportional to that value.
%\comment{[NO ME QUEDA CLARO COMO SE OBTIENEN LOS $\alpha_{ij}$ y $\beta_{ij}$.} \cefe{los alphas son el boostrap de los rewards y los beta de las visitas. La relacion alpha/beta termina estimando la media de la distribucion normal} \comment{ $v(a_i)==reward(a_i)$?]} \cefe{$v(a_i)$ es el valor que le das al nodo dependiendo de los estadisticos que venis calculando (depende de la politica de selección). En general $v(a_i)$ lo podes ver como la utilidad esperada si decidis ir por esa rama. El $reward(a_i)$ es lo que obtenes en las hojas, que se propaga de acuerdo a la funcion de backpropagation, para nosotros accuracy. En UCT si un nodo no es hoja, el reward del nodo tiene la suma de todos los $\delta$ o rewards o accuracy de las hojas debajo de el que fueron evaluadas en el dataset de training. En BTS $\alpha$ tiene la suma de los $\delta$, y para $J=1$ va a tener el mismo valor que el reward de UCT.}

\paragraph{Tree Parzen Estimator (TPE).}

TPE is a widely used decision algorithm in hyperparameter optimization \cite{bergstra2015hyperopt} that can be easily adapted to work on trees.
%
%TPE defines $p(a_i,v(a_i))$ using two densities, $l(a_i)$ and $g(a_i)$ as shown in the equation \ref{eq:tpe}.
%
%\begin{equation}\label{eq:tpe}
%    p(a_i,reward(a_i)) = \left\{\begin{matrix}
%    l(a_i) & \textup{if } y < y^* \textup{,}\\ 
%    g(a_i) & \textup{if } y \geq  y^*
%\end{matrix}\right.
%\end{equation}
%\comment{[ESTA DEFINICION NO TIENE SENTIDO. P NO ES DENSIDAD]}
%
%where $l(a_i)$ is the density formed by using the observations $\left \{a_i\right \}$ such that corresponding reward $y=reward(a_i)$ was less than $y^*$ and $g(a_i) $ is the density formed by using the remaining observations. The value of $y^*$ is chosen as a quantile $\gamma$ of the observed $y$ values and there’s no requirement for a specific model for $p(y)$ . The tree-structured form of $l$ and $g$ makes it easy to evaluate candidates based on $g(x)\textup{/}l(x)$.
%
TPE defines two densities, $l(a_i)$ and $g(a_i)$, where $l(a_i)$ is computed by using the observations $\{a_i\}$ such that corresponding reward $y=reward(a_i)$ is less than a threshold $y^*$, while $g(a_i)$ is computed using the remaining observations. The value of $y^*$ is chosen as the $\gamma$-quantile of the observed $y$ values.

To compute the functions $g(a_i)$ and $l(a_i)$, it is necessary to keep track of the rewards obtained from each node instead of accumulating them as in the case of UCT. Algorithm~\ref{alg:backprop_TPE} shows the overall process.
\begin{algorithm}
\caption{Backpropagation for TPE}
\label{alg:backprop_TPE}
\begin{algorithmic}[1]
\Function{Backpropagation}{$a$, $\Delta$}
\State $a_i \leftarrow a$ \algorithmiccomment{a is a leaf node}
\While{$a_i$ is not Null}
    \State $[reward(a_i)] \leftarrow [reward(a_i)] + [\Delta]$
%\State $a_i \leftarrow a_i$ parent  //  \textit{if $a_i$ is the root node parent is Null}
\State $a_i \leftarrow parent(a_i)$ \algorithmiccomment{if $a_i$ is the root node parent is Null}
\EndWhile
\EndFunction
\end{algorithmic}
\end{algorithm}
The tree-structured induced by the partitions used to compute $l$ and $g$ makes it easy to evaluate candidates based on the ratio $g(a_i)\textup{/}l(a_i)$, similar to the selection policy used in BTS.
%
%\begin{equation}\label{equ:yTPS}
%y^* = quantile(rewards(a_0), \gamma)
%\end{equation}
%
%where $a_0$ is the root, and it has the list of all rewards collected.
%
%\begin{equation}\label{equ:valueTPS}
%v(a_i) = \frac{g(a_i)}{l(a_i)}
%\end{equation}

%TPE is a stochastic algorithm so instead of obtaining a discrete function of value we obtain the probability of each node of being selected in the same way as with BTS through the function \ref{equ:armaxBTS}.

\section{Experiments}

To evaluate our work, we conducted experiments in two parts. In the first part, we performed an ablation study of each of the non-parametric selection policies proposed in the context of the proposed MCTS algorithm. In the second part, we compare our approach with other from the state-of-the-art.

\subsection{Experimental Setup}

The complete grammar we use in our experiments is based on the same set of components as in AutoSklearn. We employ the same hyperparameter ranges and functions as AutoSklearn when sampling and incorporates them into the grammar. For our experiments, we sample 3 values of each hyperparameter that, added to the pipeline options, extend the space to more than 183 billions possible combinations. This has to be compared with the $\sim$24K configurations of the original GramML method \cite{vazquez2022gramml}.

For evaluation, we use the OpenML-CC18 benchmark \cite{vanschoren2014openml}, a collection of 72 binary and multi-class classification datasets carefully curated from the thousands of datasets on OpenML \cite{oml-benchmarking-suites}. We use standard train-test splits, as provided by the benchmark suite. Also, to calculate validation scores (e.g. rewards), we use AutoSklearn's default strategy which consists on using 67\% of the data for training and 33\% for validation. All experiments were performed on an Amazon EC2 R5 spot instance (8 vCPU and 64GB of RAM).

\subsection{Ablation Study}

To conduct the ablation study, four tasks with low computational cost were selected from the OpenML-CC18 benchmark\footnote{These tasks, identified by task IDs 11, 49, 146819, and 10093.}. For each task, we run 100 iterations of the algorithm for each policy configuration and report the mean and standard deviation for the following metrics: time per iteration (Time Iter), number of actions per iteration (Act/Iter), simulations' repetition rate (Rep Ratio), time spent by the first iteration (Time 1st), the number of actions for the first iteration (1st Act), and the total time (Tot Time) and total number of actions (Tot Act) at the end of the 100 iterations. Time measurements only consider the time of the algorithm and do not take into account the fitting time. The simulations' repetition ratio, \ie the number of times the algorithm repeats the same paths during simulation, can be seen as a measure of the exploration efficiency of the algorithm (the lower this value, the more efficient the exploration of the search space).

Tables~\ref{tab:UCT}, \ref{tab:BTS} and \ref{tab:TPE} show results for the UTC, BTS and TPE strategies, respectively. All three methods depend on a single hyperparameter to control the overall behavior of the search algorithm, \ie the exploration-exploitation trade-off. An immediate effect of varying such parameters can be observed on the time it takes for the algorithm to complete an iteration. This has to do with the iteration ending when the algorithm reaches a leaf. If we favor exploration (i.e. increase C and J for UCT and BTS, or decrease $\gamma$ for TPE), the algorithm explores the tree in width, performs more \textit{total actions}, and takes longer to reach a leaf. On the other hand, by favoring exploitation, the algorithm performs fewer actions and reaches the terminal nodes faster. The number of actions per iteration and the time per iteration are strongly correlated.

Comparing the different strategies, we observe that UCT with $C=0.7$ is the most efficient alternative (in terms of the repetition ratio), while UCT with $C=0$ is the fastest, with the later being also the less efficient in terms of exploration. 

Finally, we consider the best version of each strategy prioritizing exploration efficiency (low iteration rate) trying to keep the time per iteration limited to $\sim$0.1. We select UCT with $C=0.7$, BTS with $J=1$, and TPE with $\gamma=0.85$ and name them $\text{GramML}^{++}_{\text{UCT}}$, $\text{GramML}^{++}_{\text{BTS}}$, and $\text{GramML}^{++}_{\text{TPE}}$, respectively.
\begin{table}
\caption{Ablation results for the UCT strategy for different values of the parameter $C$.}
\centering
\begin{tabular}{lccccc}
\toprule
 &C=0 &C=0.1 &C=0.7 &C=1 \\
\midrule
Time Iter &0,04 \scriptsize(0,01) &0,05 \scriptsize(0,01) &0,10 \scriptsize(0,02) &0,14 \scriptsize(0,04) \\
Time 1st &0,76 \scriptsize(0,31) &1,65 \scriptsize(0,49) &7,16 \scriptsize(2,50) &10,64 \scriptsize(3,99) \\
Tot Time &3,94 \scriptsize(0,64) &4,90 \scriptsize(0,54) &9,71 \scriptsize(2,14) &14,21 \scriptsize(3,84) \\
\midrule
Act/Iter &5,8 \scriptsize(0,3) &7,1 \scriptsize(0,9) &19,4 \scriptsize(4,7) &28,1 \scriptsize(7,7) \\
1st Act &165,5 \scriptsize(52,7) &324,2 \scriptsize(95,9) &1535,0 \scriptsize(545,8) &2201,2 \scriptsize(810,5) \\
Tot Act &682,5 \scriptsize(35,4) &817,2 \scriptsize(95,7) &2047,0 \scriptsize(475,9) &2914,0 \scriptsize(776,5) \\
\midrule
Rep Ratio &0,65 \scriptsize(0,04) &0,58 \scriptsize(0,07) &0,40 \scriptsize(0,08) &0,45 \scriptsize(0,10) \\
\bottomrule
\end{tabular}
\label{tab:UCT}
\end{table}
\begin{table}
\caption{Ablation results for the BTS strategy for different values of the parameter $J$.}
\centering
\begin{tabular}{lccccc}
\toprule
 &J=1 &J=10 &J=100 &J=1000 \\
\midrule
Time Iter &0,15 \scriptsize(0,05) &0,15 \scriptsize(0,06) &0,33 \scriptsize(0,23) &0,28 \scriptsize(0,12) \\
Time 1st &7,13 \scriptsize(2,19) &7,42 \scriptsize(3,86) &18,90 \scriptsize(15,42) &14,56 \scriptsize(7,05) \\
Tot Time &14,72 \scriptsize(4,51) &14,98 \scriptsize(6,27) &33,21 \scriptsize(23,20) &28,18 \scriptsize(12,40) \\
\midrule
Act/Iter &24,6 \scriptsize(7,8) &24,3 \scriptsize(11,5) &46,0 \scriptsize(30,0) &36,6 \scriptsize(15,4) \\
1st Act &1307,5 \scriptsize(398,8) &1333,2 \scriptsize(713,6) &2744,0 \scriptsize(2075,2) &2027,0 \scriptsize(927,2) \\
Tot Act &2561,2 \scriptsize(783,8) &2531,5 \scriptsize(1151,3) &4706,7 \scriptsize(3008,7) &3760,2 \scriptsize(1546,0) \\
\midrule
Rep Ratio &0,50 \scriptsize(0,13) &0,51 \scriptsize(0,10) &0,54 \scriptsize(0,09) &0,59 \scriptsize(0,04) \\
\bottomrule
\end{tabular}
\label{tab:BTS}
\end{table}
\begin{table}
\caption{Ablation results for the TPE strategy for different values of the parameter $\gamma$.}
\centering
\begin{tabular}{lccccc}
\toprule
 &$\gamma$=50 &$\gamma$=65 &$\gamma$=75 &$\gamma$=85 \\
\midrule
Time Iter &0,33 \scriptsize(0,16) &0,19 \scriptsize(0,03) &0,18 \scriptsize(0,06) &0,12 \scriptsize(0,02) \\
Time 1st &11,16 \scriptsize(7,55) &5,93 \scriptsize(1,47) &4,02 \scriptsize(2,50) &1,71 \scriptsize(0,57) \\
Tot Time &33,31 \scriptsize(15,82) &18,92 \scriptsize(3,35) &17,86 \scriptsize(6,05) &12,33 \scriptsize(1,76) \\
\midrule
Act/Iter &15,0 \scriptsize(5,5) &9,1 \scriptsize(1,3) &7,8 \scriptsize(1,8) &5,9 \scriptsize(0,2) \\
1st Act &740,2 \scriptsize(382,0) &434,7 \scriptsize(81,8) &286,5 \scriptsize(128,4) &143,0 \scriptsize(25,7) \\
Tot Act &1605,5 \scriptsize(555,8) &1018,7 \scriptsize(129,8) &882,5 \scriptsize(187,2) &697,0 \scriptsize(23,0) \\
\midrule
Rep Ratio &0,47 \scriptsize(0,06) &0,59 \scriptsize(0,06) &0,60 \scriptsize(0,05) &0,68 \scriptsize(0,01) \\
\bottomrule
\end{tabular}
\label{tab:TPE}
\end{table}
\subsection{Comparison with other techniques}

In this section, we compare our methods with others from the literature. For each task, each method is run for an hour (time budget). We report performance on the test set proposed in the benchmark suite. We compare the different variations of our approach to AutoSklearn \cite{feurer2015efficient} and MOSAIC \cite{rakotoarison2018automl} as they allow us to rely on the exact same set of basic ML components. In all cases, we set a maximum fitting time of 300 seconds, as the time limit for a single call to the machine learning model. If the algorithm runs beyond this time, the fitting is terminated. In our case, we return a reward of zero for methods exceeding this time limit. For each task, we ranked the performance of all systems and reported the average ranking (lower is better) and average performance score (higher is better) calculated as 1 - regret (the difference between maximum testing performance found so far by each method and the true maximum) at each time step.

\begin{figure}
     \centering
     \begin{subfigure}[b]{0.48\textwidth}
         \centering
         \includegraphics[width=\textwidth]{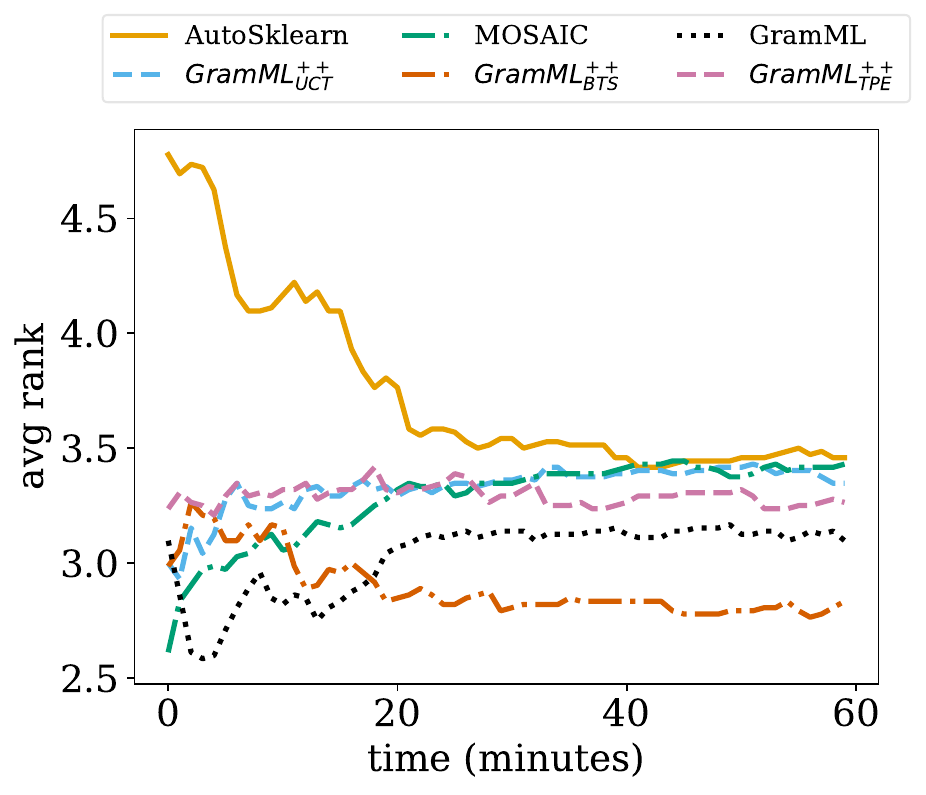}
         \caption{Average rank over 1 hour execution.}
         \label{fig:avg_rank}
     \end{subfigure}
     \hfill
     \begin{subfigure}[b]{0.49\textwidth}
         \centering
         \includegraphics[width=\textwidth]{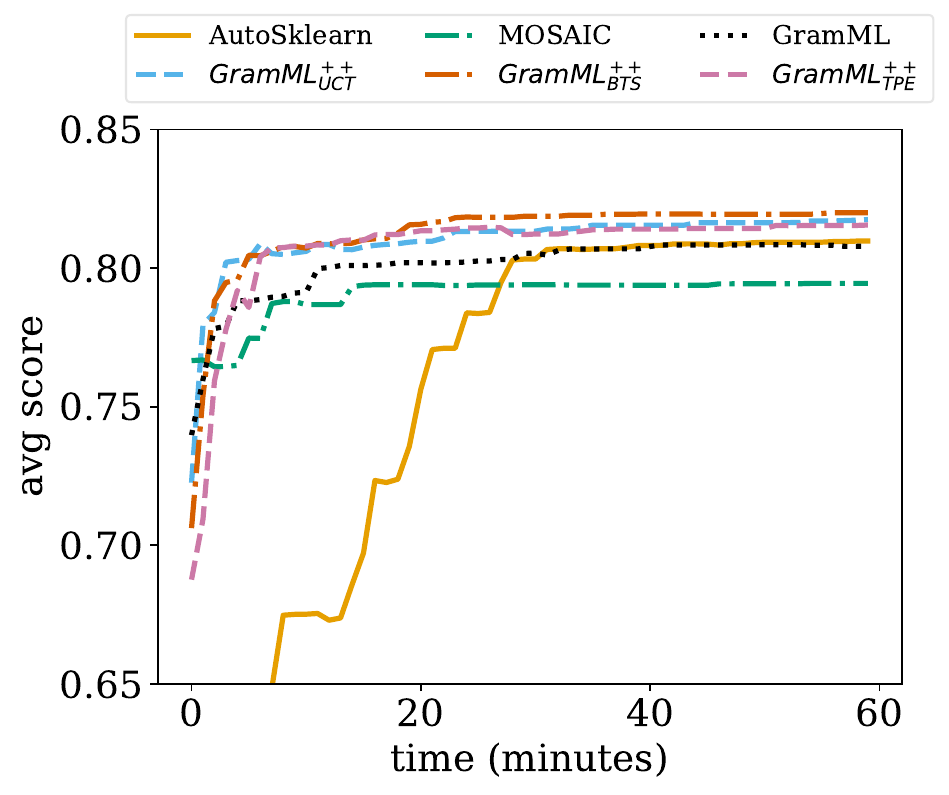}
         \caption{Average score over 1 hour execution.}
         \label{fig:avg_score}
     \end{subfigure}
    \caption{Results of the approaches on OpenML-CC18 benchmark.}
    \label{fig:Result comparison}
\end{figure}

Comparative results are shown in Figure~\ref{fig:Result comparison}. Figure~\ref{fig:avg_rank} shows the average rank across time. A marked difference of $\text{GramML}^{++}_{\text{BTS}}$ variant can be seen over MOSAIC, AutoSklearn and GramML. In particular, $\text{GramML}^{++}_{\text{BTS}}$ have the best ranking over time, although it is surpassed in the first few minutes by MOSAIC and GramML. There also seems to be a slight decreasing trend for $\text{GramML}^{++}_{\text{BTS}}$ which would seem to indicate that the results could improve with more time.

In addition, Figure~\ref{fig:avg_score} shows the average score across time. $\text{GramML}^{++}$ variants have a visible difference from MOSAIC, AutoSKLearn and GramML. The best average score is also achieved by $\text{GramML}^{++}_{\text{BTS}}$. An interesting observation is that the previous version of GramML was eventually slightly outperformed in terms of average score by AutoSKLearn. One reason for this was that GramML did not utilize hyperparameters. Conversely, the variants of $\text{GramML}^{++}$ demonstrate a significant improvement over their predecessor and AutoSKLearn. It should be noted that the average regret measure may not be very representative if the accuracy varies greatly between tasks. However, we believe that the graph provides important information together with the average ranking for the comparison between techniques in the OpenML-CC18 benchmark.
%
%GramML compared to MOSAIC and AutoSklearn\footnote{We use AutoSklearn v\texttt{0.7.0} and MOSAIC v\texttt{0.1b0}, both of which rely on scikit-learn v\texttt{0.22.2.post1}}. During the first few minutes, MOSAIC shows the best performance among the three systems but it becomes surpassed by GramML shortly after. An interesting phenomenon can be observed around minute 20, where the performance of MOSAIC and AutoSklearn crosses. This might be attributed to the time it takes for the surrogate model in AutoSklean to stabilize. After this point, the performance of AutoSklearn improves over MOSAIC steadily. Interestingly, BruteForce seems to be a strong baseline and its performance is similar to that of AutoSklearn.%

\begin{figure*}
        \centering
        \begin{subfigure}[b]{0.49\textwidth}
            \centering
            \includegraphics[width=\textwidth]{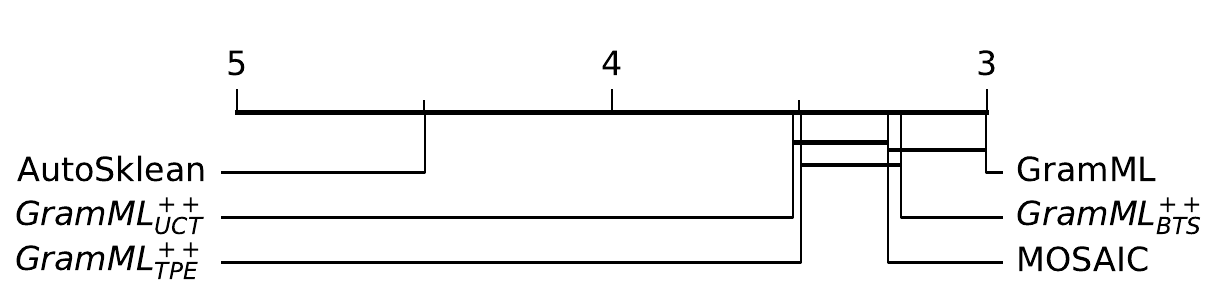}
            \caption{15 minutes}    
            \label{fig:15min}
        \end{subfigure}
        \hfill
        \begin{subfigure}[b]{0.49\textwidth}  
            \centering 
            \includegraphics[width=\textwidth]{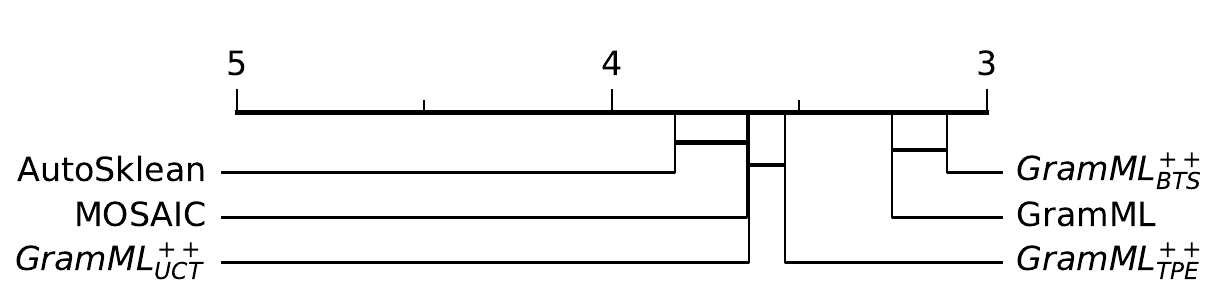}
            \caption{30 minutes}   
            \label{fig:30min}
        \end{subfigure}
        \vskip\baselineskip
        \begin{subfigure}[b]{0.49\textwidth}   
            \centering 
            \includegraphics[width=\textwidth]{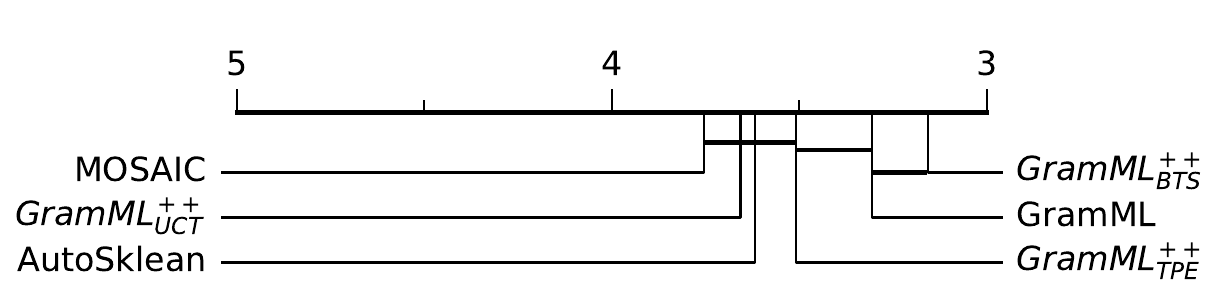}
            \caption{45 minutes}   
            \label{fig:45min}
        \end{subfigure}
        \hfill
        \begin{subfigure}[b]{0.49\textwidth}   
            \centering 
            \includegraphics[width=\textwidth]{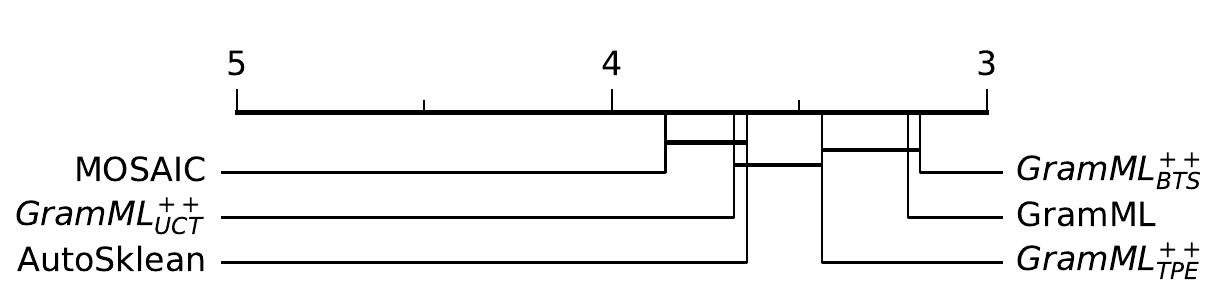}
            \caption{1 hour}    
            \label{fig:1hour}
        \end{subfigure}
        \caption{CD plots with Nimenyi post-hoc test} 
        \label{fig:cdplots}
\end{figure*}

In order to check the presence of statistically significant differences in the average rank distributions, Figure \ref{fig:cdplots} shows the results using critical difference (CD) diagrams \cite{demvsar2006statistical} at 15, 30, 45 and 60 minutes. We use a non-parametric Friedman test at p $<$ 0.05 and a Nemenyi post-hoc test to find which pairs differ \cite{gijsbers2022amlb}. If the difference is significant, the vertical lines for the different approaches appear as well separated. If the difference is not significant, they are joined by a thick horizontal line. 

The diagrams show statistically significant differences at 15 minutes between AutoSKLearn and 
the other variants. At 30 minutes both $\text{GramML}^{++}_{\text{BTS}}$ and GramML have significant difference with MOSAIC and AutoSKLearn. At minute 45, the difference between $\text{GramML}^{++}_{\text{BTS}}$ and the other techniques remains, but there is no significant difference between the previous version of GramML and $\text{GramML}^{++}_{\text{TPE}}$. Finally, at 60 minutes, $\text{GramML}^{++}{\text{BTS}}$ showed a significant difference compared to MOSAIC and AutoSKLearn, while $\text{GramML}^{++}{\text{TPE}}$ showed a significant difference only compared to MOSAIC. However, $\text{GramML}^{++}_{\text{UCT}}$ did not show a statistically significant difference compared to either MOSAIC or AutoSKLearn.
%
%Something to keep in mind is that these diagrams show only a photo of the techniques at the given minute. If we take into account a window of 15 minutes instead of just an instant, the MCTS variants have significant differences to all other techniques in all minutes.
%
\section{Conclusions and Future Work}

This article presents a model-free approach for grammar-based AutoML that integrates hyperparameter search as an extension to GramML \cite{vazquez2022gramml}. The approach involves two steps: (1) incorporating hyperparameter values into the grammar using grammar rules and (2) modifying the Monte Carlo Tree Search (MCTS) algorithm to improve the collection of the best pipeline configuration, implement pruning, and support different selection policies and backpropagation functions. The evaluation involved an ablation study to assess the efficiency of each selection policy and a comparison with state-of-the-art techniques, which showed significant improvements, particularly with a variant that uses bootstrapped Thompson Sampling \cite{eckles2014thompson}. Consecuently, this work demonstrates the effectiveness of incorporating hyperparameter search into grammar-based AutoML, and provides a promising approach for addressing the challenges of larger search spaces.

Furthermore, this work opens up several avenues for future research. In model-based AutoML, meta-learning can improve exploration efficiency \cite{vanschoren2019meta}. However, the application of meta-learning in model-free AutoML based on grammars remains an open question. Additionally, incorporating resource information into the AutoML objective function is also an important direction to follow \cite{vazquez2023general}. Finally, the speed of the search algorithms is an important factor that can be improved through horizontal scalability \cite{bourki2011scalability}. Therefore, future research should aim to parallelize the algorithm for more efficient resource utilization.

\bibliographystyle{splncs04}
\bibliography{references}

\end{document}